\documentclass{trbunofficial}
\usepackage{graphicx}

\usepackage[hidelinks]{hyperref}

\usepackage{booktabs}
\usepackage{subcaption}
\usepackage{multirow}

\usepackage{amsfonts}

\AuthorHeaders{Zhu et al.}
\title{TransFollower: Long-Sequence Car-Following Trajectory Prediction through Transformer}

\author{%
  \textbf{Meixin Zhu}\\
  Research Assistant\\
  Department of Civil and Environmental Engineering\\
 University of Washington\\
  meixin92@uw.edu\\
  
\hfill\break%
\textbf{Simon S. Du, Ph.D.}\\
Assistant Professor \\
Paul G. Allen School of Computer Science \& Engineering\\
University of Washington\\
ssdu@cs.washington.edu\\

\hfill\break%
  \textbf{Xuesong Wang, Ph.D.}\\
  Professor\\
School of Transportation Engineering\\
Tongji University\\
wangxs@tongji.edu.cn\\
\hfill\break
  \textbf{Hao (Frank) Yang}\\
  Research Assistant\\
  Department of Civil and Environmental Engineering\\
 University of Washington\\
  haoya@uw.edu\\
\hfill\break
  \textbf{Ziyuan Pu}\\
  Assistant Professor\\
  School of Engineering\\
 Monash University\\
  ziyuan.pu@monash.edu\\
  \hfill\break
  \textbf{Yinhai Wang, Ph.D., Corresponding Author}\\
  Professor\\
  Department of Civil and Environmental Engineering\\
 University of Washington\\
  yinhai@uw.edu
}

\begin{document}
\maketitle
\section{Abstract}

Car-following refers to a control process in which the following vehicle (FV) tries to keep a safe distance between itself and the lead vehicle (LV) by adjusting its acceleration in response to the actions of the vehicle ahead. The corresponding car-following models, which describe how one vehicle follows another vehicle in the traffic flow, form the cornerstone for microscopic traffic simulation and intelligent vehicle development. One major motivation of car-following models is to replicate human drivers' longitudinal driving trajectories. To model the long-term dependency of future actions on historical driving situations, we developed a long-sequence car-following trajectory prediction model based on the attention-based Transformer model. The model follows a general format of encoder-decoder architecture. The encoder takes historical speed and spacing data as inputs and forms a mixed representation of historical driving context using multi-head self-attention. The decoder takes the future LV speed profile as input and outputs the predicted future FV speed profile in a generative way (instead of an auto-regressive way, avoiding compounding errors). Through cross-attention between encoder and decoder, the decoder learns to build a connection between historical driving and future LV speed, based on which a prediction of future FV speed can be obtained. We train and test our model with 112,597 real-world car-following events extracted from the Shanghai Naturalistic Driving Study (SH-NDS). Results show that the model outperforms the traditional intelligent driver model (IDM), a fully connected neural network model, and a long short-term memory (LSTM) based model in terms of long-sequence trajectory prediction accuracy. We also visualized the self-attention and cross-attention heatmaps to explain how the model derives its predictions. 

\hfill\break%
\noindent\textit{Keywords}: Intelligent transportation, electric vehicles, anomaly detection, connected vehicle, crash recognition
\newpage

\section{Introduction}\label{I}
Car-following is the most common driving task. It refers to a process where the following vehicle (FV) tries to keep a safe distance between itself and the lead vehicle (LV) by adjusting its acceleration in response to the actions of the vehicle ahead \cite{chakroborty1999evaluation}. The corresponding car-following models are functions that determine FV's future accelerations based on current (and historical) driving situations. Car-following models are the cornerstone for microscopic traffic simulation and intelligent vehicle development \cite{brackstone1999car}. 

Specifically, in the modeling of car-following behavior, the input at a certain time step $t$ is described by $[s_t, v_t, \Delta v_t]$, where $s$ is the inter-vehicle spacing, $v$ is the speed of an FV, and $\Delta v$ is the relative speed between a lead and following vehicle. The model's output is the predicted FV speed at the next time step $\Tilde{v}_{t + 1} = f([s_t, v_t, \Delta v_t])$. The goal is to minimize the mean squared errors between the predicted $\Tilde{v}_{t + 1}$ and observed $v_{t + 1}$ future speeds. 

Currently, we have two types of car-following models that solve the aforementioned problem: 1) traditional car-following models with explicit mathematical equations and 2) data-driven models that fit a machine learning model using car-following trajectory data (see the following section for a detailed review). However, existing methods from both types of models have one or more of the following limitations:
\begin{enumerate}
    \item \textbf{Strong Inductive Bias.} This is more obvious with traditional car-following models in that they have fixed forms of equations specifying how the accelerations of an FV will depend on relative speeds and spacing. For example, the intelligent driver model (IDM) assumes that drivers have a desired following speed and gap that can be calculated based on some equations. But these equations might not be correct for real-world driving behavior modeling. 
    \item \textbf{Ignoring Historical Driving Context.} Most existing car-following models assume the future FV action will only depend on the current instantaneous car-following state. However, with a given current car-following state, we can have multiple historical driving situations that lead to this current state, each of which will cause different FV reactions. Also, considering that human drivers do take information from the last several seconds to make decisions, it would be straightforward to take historical driving context into account for driving behavior modeling.
    \item \textbf{Lacking Long-Sequence Prediction Capability.} Most existing car-following models only predict one step forward for the FV trajectory. This has no problem if we can constantly query for the true car-following state before prediction. However, in the testing phase, this is impossible since we do not have access to the true car-following state. One way to do multi-step prediction using single-step prediction models is using an auto-regressive prediction mechanism, which means we make a next-step prediction based on the predicted last-step state. The issue with this approach is that errors accumulated across multiple steps will make the final predictions deviate far away from the real ones. 
    
\end{enumerate}

To address these limitations, we propose a long-sequence car-following trajectory prediction model that is data-driven; captures the temporal dependencies of future driving decisions on historical driving context; and can achieve fast and accurate prediction of long-sequence car-following trajectories. The proposed model is built upon the attention-based Transformer model \cite{vaswani2017attention}, which is a state-of-the-art method for long-term dependency modeling in natural language processing (NLP) \cite{devlin2018bert, brown2020language} and computer vision (CV) \cite{liu2021swin}. The model follows a general format of encoder-decoder architecture. The encoder takes historical speed and spacing data as inputs and forms a mixed representation of historical driving context using multi-head self-attention. The decoder takes the future LV speed profile as input and outputs the predicted future FV speed profile in a generative way (instead of an auto-regressive way, avoiding compounding errors). Through cross-attention between encoder and decoder, the decoder learns to build a connection between historical driving and future LV speed, based on which a prediction of FV speed can be obtained. We train and test our model with 112,597 real-world car-following events extracted from the Shanghai Naturalistic Driving Study (SH-NDS). Results show that the model outperforms the traditional IDM, a fully connected neural network model, and a long short-term memory (LSTM) based model by a large margin in long-sequence trajectory prediction accuracy. We also visualized the self-attention and cross-attention heatmaps to explain how the model derives its predictions. 

The contributions of this study are:
\begin{itemize}
    \item Proposed an innovative car-following modeling architecture based on the encoder-decoder framework.
    \item Addressed long-term historical dependency and long-sequence trajectory prediction in driving behavior modeling using attention-based models. 
    \item Provided some visualization explorations for learning-based behavioral models understanding.
\end{itemize}





\section{Related Work}
\subsection{Car Following}
The Gazis-Herman-Rothery (GHR) car-following model \cite{chandler1958traffic, gazis1961nonlinear} was developed by researchers at the General Motors Research Laboratories in the middle of the 1950s, and it was the first study of car-following behaviors and models. The GHR model assumes that the FV decides its acceleration rate based on cues such as the relative speed between itself and the LV as a stimulus-response model. Subsequently, lots of car-following models have emerged. The most notable ones include Helly's model \cite{helly1959simulation}, Gipps model \cite{gipps1981behavioural}, Wiedemann model \cite{wiedemann1974simulation}, IDM \cite{treiber2000congested}, and optimal velocity model \cite{bando1995dynamical}. According to Zhu et al. \cite{zhu2018modeling}, the IDM has the best behavior prediction performance among traditional car-following models. For a detailed review of traditional car-following models, readers can refer to Brackstone and McDonald \cite{brackstone1999car} and Saifuzzaman and Zheng \cite{saifuzzaman2014incorporating}.

\subsection{Data-Driven Car-Following Models}
The current availability of high-fidelity traffic data, as well as the resulting data-driven methods, have opened the door to directly modeling drivers' car-following behavior from massive field data. Data-driven techniques are more flexible than traditional models because they allow for the inclusion of extra characteristics that affect driving behavior, resulting in richer models. Based on the machine models used, previous data-driven car-following models can be categorized into the following types:
\begin{itemize}
    \item \textbf{Nonparametric}. He et al. \cite{he2015simple} proposed a simple nonparametric car-following model with k-nearest neighbors, which outputs the average of the most similar cases, i.e., the most likely driving behavior under the current circumstance. Papathanasopoulou and Antoniou \cite{PAPATHANASOPOULOU2015496} developed a nonparametric data-driven car-following model based on locally weighted regression, the Loess model.
    \item \textbf{Fully connected neural networks}. Jia et al. \cite{hongfei2003develop} introduced a four-layer neural network (including one input layer, two hidden layers, and one output layer). This neural network takes relative speed, desired speed, follower speed, and gap distance as inputs, using them to predict follower acceleration.
    \item \textbf{Recurrent neural networks (RNN)}. Zhou et al. \cite{zhou2017recurrent} used vanilla RNN to model drivers' car-following behavior, with a focus on predicting traffic oscillations. Wang et al. \cite{wang2017capturing} used a variant of RNN called Gated Recurrent Unit (GRU) \cite{cho2014learning} to model car-following behaviors. Ma and Qu \cite{ma2020sequence} proposed a sequence-to-sequence car following prediction model using another variant of RNN called LSTM network \cite{hochreiter1997long}. 
    \item \textbf{Other methods}. There are studies that used deep reinforcement learning (RL) \cite{zhu2018human}, inverse RL \cite{gao2018car}, and generative adversarial imitation learning \cite{zhou2020modeling} to model car-following behaviors. For a more detailed review of recent car-following studies, please refer to Li et al. \cite{li2020trajectory}.
\end{itemize}

\section{Background}
\subsection{Sequence-to-Sequence (Seq2Seq) Model}
Seq2Seq architecture \cite{sutskever2014sequence} is designed for machine learning problems with sequences as both input and output. An encoder, an intermediate vector, and a decoder are the three components of a Seq2Seq model. A stack of recurrent units or Transformer blocks is used to create an encoder. The encoder vector, also known as the context vector, is the encoder's final hidden state, which encodes all of the information from the input data. The encoder vector is then used as the initial hidden state of the decoder, which is likewise made up of a stack of Transformer blocks or recurrent units. The decoder finally generates a sequence of output vectors.

Mathematically, given an input sequence $\mathbf{X} = (\mathbf{x_1}, \mathbf{x_2}, \dots, \mathbf{x_m})$ with $m$ being input sequence length and $\mathbf{x_i} \in \mathbb{R}^{d}$, the encoder maps $\mathbf{X}$ into context vector $\mathbf{z}$. Given $\mathbf{z}$, the decoder then generate a sequence $(\mathbf{y_1}, \mathbf{y_2}, \dots, \mathbf{y_n})$ of output vectors with $n$ being output sequence length and $\mathbf{y_j} \in \mathbb{R}^o$. It can happen that $m \neq n$ and $d \neq o$.

\subsection{Transformer}
Transformer \cite{vaswani2017attention} follows a similar architecture as Seq2Seq models, but is based solely on attention mechanisms. It contains an encoder and a decoder, both of them formed by a stack of transformer blocks. Embedding layers are also used in both encoder and decoder to transform the original data inputs to vectors $\in \mathbb{R}^{d_{model}}$.

\subsubsection{Encoder}
The encoder is made up of $N$ identical blocks stacked on top of each other. Each block is divided into two parts: a multi-head self-attention mechanism, and a position-wise fully connected feed-forward network. Residual connections and layer normalization are added for both parts. In mathematical description, a transformer block is a parameterized function class $f_{\theta}: \mathbb{R}^{p \times d} \rightarrow \mathbb{R}^{p \times d}$ where $p$ is the sequence length for both input and output, and $d=d_{model}$ is the dimension of  the model. 


If $\mathbf{X},\mathbf{Z}  \in \mathbb{R}^{p \times d}$ are the input and output sequences for the transformer block, respectively, and $\mathbf{x_i}, \mathbf{z_i} \in \mathbb{R}^d$ are the $i^{th}$ element of $\mathbf{X}$, and $\mathbf{Z}$, then the derivation of $\mathbf{z_i}$ from $\mathbf{x_i}$ can be described using the following equations \cite{thickstuntransformer}:

\begin{equation}\label{eq:1}
    Q^{(h)}\left(\mathbf{x}_{i}\right)=W_{h, q}^{T} \mathbf{x}_{i}, \quad K^{(h)}\left(\mathbf{x}_{i}\right)=W_{h, k}^{T} \mathbf{x}_{i}, \quad V^{(h)}\left(\mathbf{x}_{i}\right)=W_{h, v}^{T} \mathbf{x}_{i}, \quad W_{h, q}, W_{h, k}, W_{h, v} \in \mathbb{R}^{d \times k},
\end{equation}
\begin{equation}\label{eq:2}
    \alpha_{i, j}^{(h)}=\operatorname{softmax}_{j}\left(\frac{\left\langle Q^{(h)}\left(\mathbf{x}_{i}\right), K^{(h)}\left(\mathbf{x}_{j}\right)\right\rangle}{\sqrt{k}}\right),
\end{equation}
\begin{equation}\label{eq:3}
    \mathbf{u}_{i}^{\prime}=\sum_{h=1}^{H} W_{c, h}^{T} \sum_{j=1}^{p} \alpha_{i, j}^{(h)} V^{(h)}\left(\mathbf{x}_{j}\right), \quad W_{c, h} \in \mathbb{R}^{k \times d},
\end{equation}
\begin{equation}\label{eq:4}
    \mathbf{u}_{i}=\operatorname{LayerNorm}\left(\mathbf{x}_{i}+\mathbf{u}_{i}^{\prime} ; \gamma_{1}, \beta_{1}\right), \quad \gamma_{1}, \beta_{1} \in \mathbb{R},
\end{equation}
\begin{equation}\label{eq:5}
    \mathbf{z}_{i}^{\prime}=W_{2}^{T} \operatorname{ReLU}\left(W_{1}^{T} \mathbf{u}_{i}\right), \quad W_{1} \in \mathbb{R}^{d \times m}, W_{2} \in \mathbb{R}^{m \times d},
\end{equation}
\begin{equation}\label{eq:6}
    \mathbf{z}_{i}=\operatorname{LayerNorm}\left(\mathbf{u}_{i}+\mathbf{z}_{i}^{\prime} ; \gamma_{2}, \beta_{2}\right), \quad \gamma_{2}, \beta_{2} \in \mathbb{R},
\end{equation}
where all the $W$s are weight matrices for the model, $Q^{(h)}, K^{(h)}, V^{(h)}$ are query, key, and value functions for attention head $h$ (among a total $H$ attention heads), respectively; $k$ is the dimension of an attention head; $m$ is the dimension of the linear layer; $\gamma_{1}, \beta_{1},\gamma_{2}, \beta_{2}$ are learnable parameters for layer normalization \cite{ba2016layer}; $\alpha_{i, j}^{(h)}$ is normalized attention weight between $\mathbf{x}_i$ and $\mathbf{x}_j$; softmax$_j$ indicates we take the softmax normalization over the $d$-dimensional vector indexed by $j$; $\mathbf{u}_{i}^{\prime}$ is the results of multi-head attention for $\mathbf{x}_i$; $\mathbf{u}_{i}$ is the result of residual adding of $\mathbf{x}_i$ and $\mathbf{u}_{i}^{\prime}$ followed by layer normalization; $\mathbf{z}_{i}^{\prime}$ is the result of point-wise fully connected feed-forward layer for $\mathbf{u}_{i}$; and $\mathbf{z}_{i}$ is the final result generated from residual adding of $\mathbf{z}_{i}^{\prime}$ and $\mathbf{u}_{i}$ followed by layer normalization. 

Equations \ref{eq:1} to \ref{eq:3} describe the process of multi-head self attention. In Equation \ref{eq:1}, $\mathbf{x}_i$ is mapped into $k$-dimensional query, key, and value vectors. In Equation \ref{eq:2}, the attention weight between position $i$ and $j$ is calculated based the inner product query vector of $i$ and key vector of $j$. In Equation \ref{eq:3}, the self-attention result of $\mathbf{x}_i$ is calculated as summing up the value vectors of all locations, weighted by the attention weights from Equation \ref{eq:2}. The weighted-sum vectors for all attention heads are then transformed from $k$ dimension to $d$ dimension and added up. Equations \ref{eq:4} and \ref{eq:5} describe the process of fully connected feed-forward network calculation. 

\subsubsection{Decoder}
The decoder also consists of a stack of $N$ identical Transformer Blocks. What is different from the encoder part is that the decoder has an additional layer that performs multi-head cross-attention over the output from the last layer of the encoder. For this cross-attention layer, the query vectors are from the decoder but the key and value vectors are from the encoder. In this way, the decoder selectively combines information from the encoder results. 

\section{Proposed Method}
\begin{figure}[!h]
\centering
\includegraphics[width=0.8\linewidth]{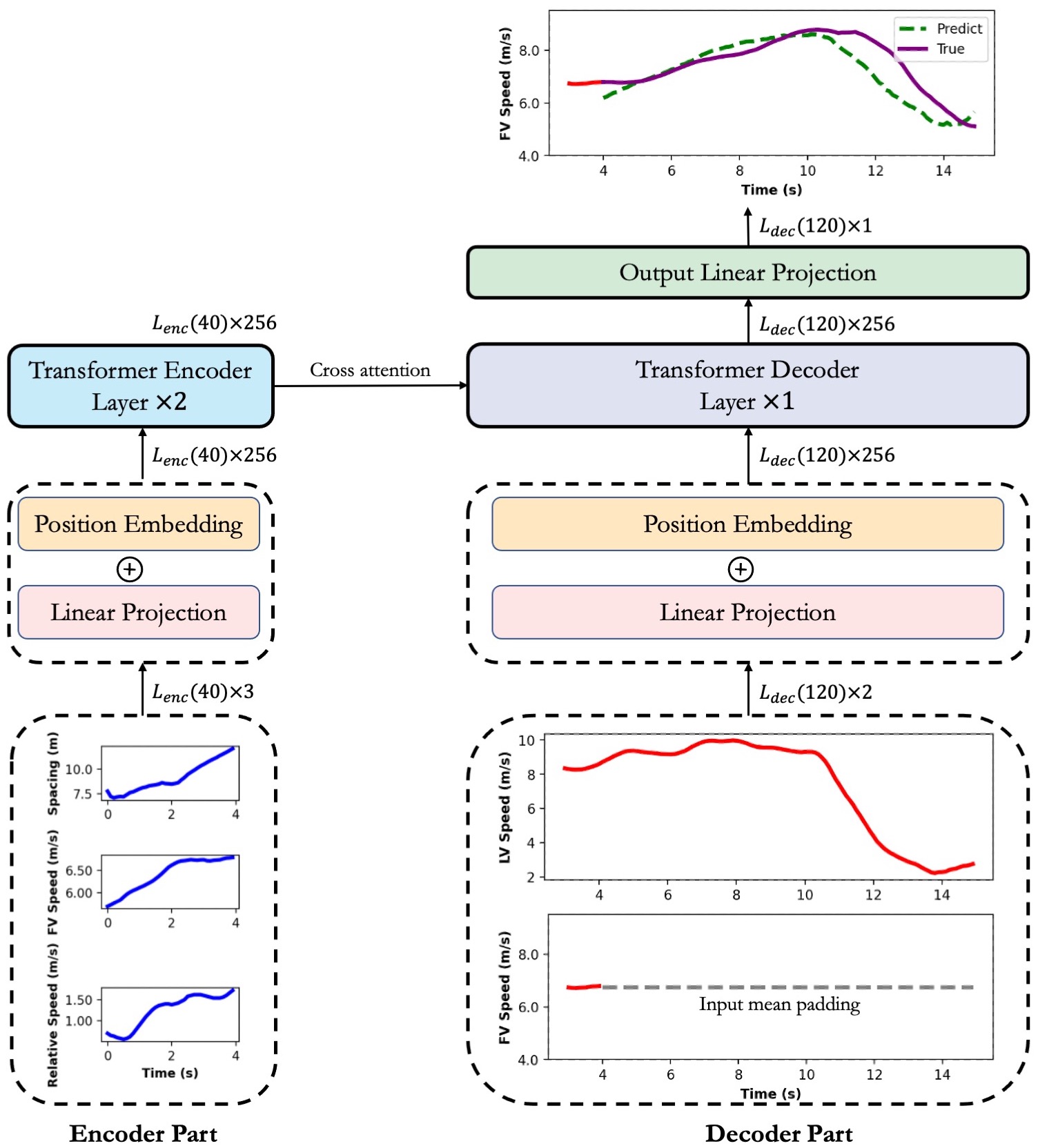}
\caption{\textbf{Transformer-based car-following model.} The encoder takes historical speed and spacing data as inputs, forming a mixed representation of historical driving context using multi-head self-attention. The decoder takes future LV speed profile and past FV speed profile concatenated with future prediction placeholder as inputs, generating FV speed predictions in a generative style. In this figure's example, we are given past data from 0 to 4 seconds, future LV speed data from 4 to 15 seconds. The objective is to predict future FV speed data from 4 to 15 seconds.}
\label{fig:model}
\end{figure}

The proposed Transformer network for car-following modeling is illustrated in Fig. \ref{fig:model}. The model architecture includes three parts: data pre-processing, encoder, and decoder. The encoder is for processing historical driving data, and the decoder is for processing the future speed profile of the LV and generating prediction results. 

\subsection{Data Pre-processing}
The data pre-processing part includes 1) a linear layer that maps the low-dimension car-following state vectors to high-dimension vectors that match the model dimension of the transformer model (set as 256 in our study), and 2) a position encoding layer that maps input time steps into learnable continuous vectors. Positional encoding is a general component for Transformer-based models since pure attention modules process all inputs in parallel, ignoring the sequencing of inputs. In this study, we used a learnable lookup table that stores embeddings of time step positions from 1 to 150. 

\subsection{Encoder}
The encoder in the TransFollower model processes historical speed and spacing data and forms a mixed representation of historical driving context using multi-head self-attention. At time step $t$, assume we consider $H$ historical steps' information, then the inputs to the model are: $\mathbf{x}_{t-(H-1)}$, $\dots$, $\mathbf{x}_{t-1}$, $\mathbf{x}_t$ where $\mathbf{x}_t = [s_t, v_t, \Delta v_t]$ is the car-following state at time step $t$, as described in the Introduction Section. As shown in Fig. \ref{fig:model}, we use the previous 4-second of data as the historical context. Since the data sampling rate is 10 Hz in our example, the raw inputs to the encoder $\in \mathbb{R}^{40 \times 3}$ where 40 is the sequence length and 3 is the number of dimensions for car-following states. After passing through the data pre-processing part, the raw inputs are transformed into hidden vectors in the shape of $\mathbb{R}^{40 \times 256}$. The hidden vectors then go through a stack of Transformer encoder layers (2 layers in our study) for multi-head self-attention representation learning. The final outputs of the encoder are in the shape of $\mathbb{R}^{40 \times 256}$, which will be used for cross attention in the decoder part. For the Transformer encoder layer, we set the number of heads as 8, the feedforward layer dimension as 1024, and the dropout probability as 0.1. 

\subsection{Decoder}
The decoder takes the future speed profile of LV as input and outputs the predicted future FV speed profile in a generative way \cite{zhou2021informer} (instead of an auto-regressive way, avoiding compounding error). Assume we use $D$ time steps of historical data as an initial token for the decoder, and we want to predict the trajectory of an FV for the future $T$ time steps. Then the raw inputs to the decoder can be represented as:
\begin{equation}\label{eq:dec_input}
\begin{split}
&\mathbf{X}_{\text {dec }}= \left[\mathbf{v}_{t-D:t+T}^{LV}, \mathbf{v}^{FV}\right]\\
&\mathbf{v}^{FV} = \operatorname{Concat}\left(\mathbf{v}_{t-D:t}^{FV}, \mathbf{v}_{\text{mean}}^{FV}\right) 
\end{split}
\end{equation}
where $t$ is the current time step; $\mathbf{v}_{t-D:t+T}^{LV} \in \mathbb{R}^{D+T}$ is the speed data of LV from $t-D$ to $t+T$; $\mathbf{v}_{t-D:t}^{FV} \in \mathbb{R}^{D}$ is the historical $D$ time steps' FV speed data; and $\mathbf{v}_{\text{mean}}^{FV} \in \mathbb{R}^{T}$ is the placeholder for future FV speed filled with mean of $\mathbf{v}_{t-D:t}^{FV}$. 

In this study, we set $D = 10, T = 110$. In this case, $\mathbf{X}_{\text {dec }}$ is in the shape of $\mathbb{R}^{(110+10) \times 2}$. Similar to the encoder part, the raw inputs of the decoder will go through the data pre-processing layer and one layer of Transformer decoder layer. In the Transformer decoder layer, both self-attention among the decoder inputs and cross-attention between decoder inputs and encoder outputs will be conducted. This allows the model to condition the future prediction on both future LV speed and past driving context. The outputs of the Trasnformer decoder layer ($\in \mathbb{R}^{120 \times 256}$) will go through a linear projection layer, generating a vector $\in \mathbb{R}^{120 \times 1}$ as the prediction of future FV speed. Given the initial car-following state, the observed LV speed profile, and the predicted FV speed profile, the predicted spacing between two vehicles can be inferred according to the following system update equations \cite{zhu2020safe}:

\begin{flalign}\label{eq:update}
\begin{split}
&\Delta V(t+1)=V_{LV}(t+1)-V_{FV}(t+1)\\
&S(t+1)=S(t)+\frac{\Delta V(t)+\Delta V(t+1)}{2}*\Delta T
\end{split}
\end{flalign}
where $\Delta T$ is the simulation time interval, set as $0.1s$ in this study, $S$ represents the spacing between two vehicles, and $V_{FV}, V_{LV}$ are the velocity of the FV and LV, respectively. 

By comparing the predicted and observed spacing and FV speed, a loss value can be calculated by summing up the Mean Squared Errors (MSEs) of spacing and FV speed. When calculating the MSE loss, only \textbf{future} spacing and speed predictions are considered (meaning the first $D$ time steps of the outputs are ignored). The loss is then backpropagated from the decoder prediction to the entire model for optimization. 



\section{Data and Experiments}
\subsection{Shanghai Naturalistic Driving Study (SH-NDS)}
Real-world car-following events from the Shanghai Naturalistic Driving Study (SH-NDS) were used to train and test the model. The SH-NDS was jointly conducted by Tongji University, General Motors, and the Virginia Tech Transportation Institute, with the aims of better understanding the vehicle usage, vehicle operation, and safety consciousness of Chinese drivers. The study used 5 passenger vehicles equipped with the second Strategic Highway Research Program (SHRP 2) DAS \cite{dingus2015naturalistic} to collect real-world driving data, starting in December 2012 and ending in December 2015. Each participant drove the vehicle for two months, and a total of 60 licensed Shanghai drivers' data were collected, with the cumulative mileage being 161,055 km.

\begin{figure}[!h]
\centering
\includegraphics[width=0.8\linewidth]{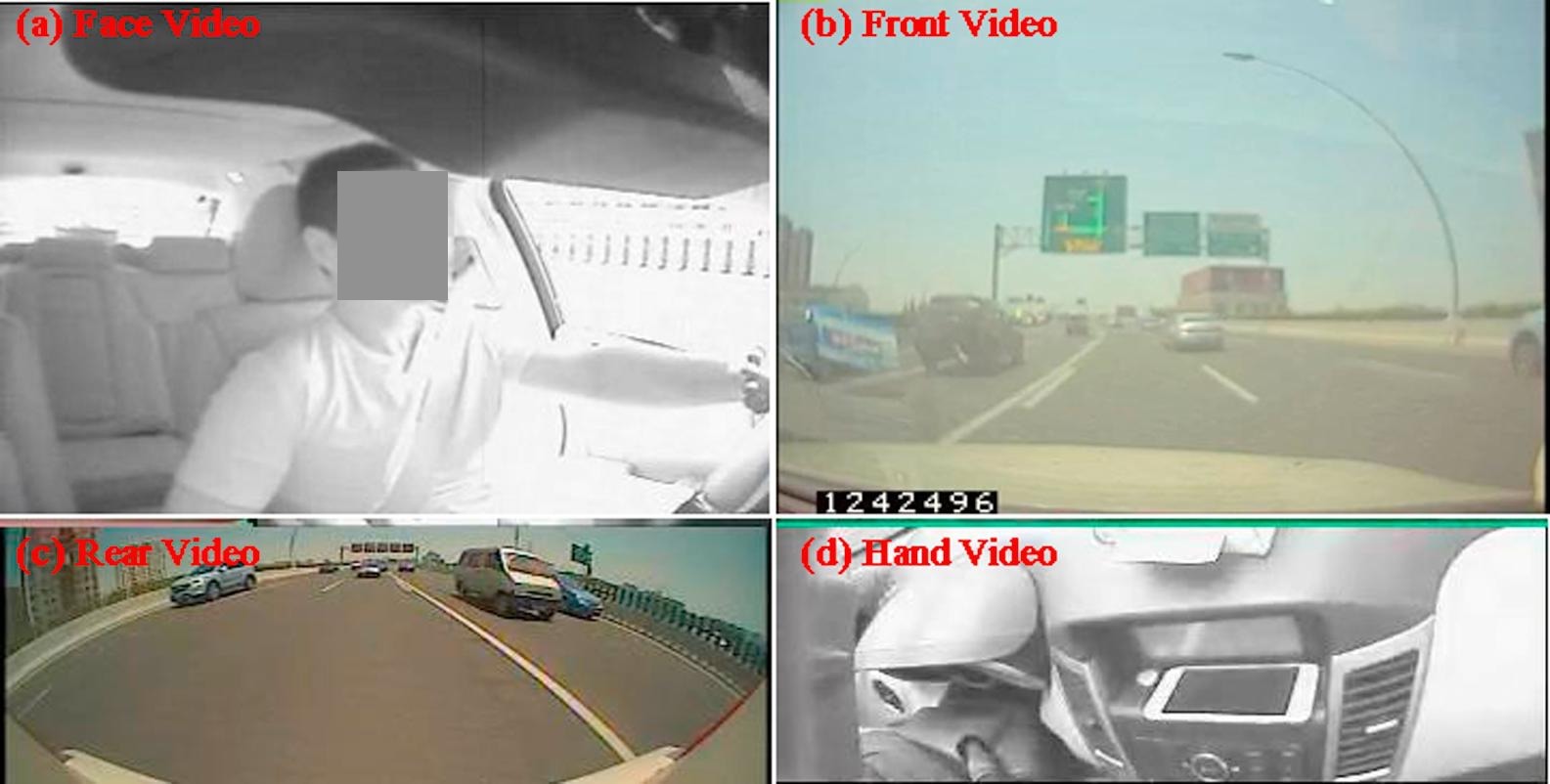}
\caption{SH-NDS four camera views \cite{zhu2018human}.}
\label{fig:shnds}
\end{figure}

The DAS consists of the following parts: a forward radar that measures distance and relative speed to vehicles ahead, an accelerometer measuring longitudinal and lateral acceleration, a GPS sensor, an interface box to collect vehicle CAN Bus data, and four synchronized video cameras that monitor the driver's face, the forward roadway, the roadway behind the vehicle, and the driver's hand maneuvers, as shown in Figure \ref{fig:shnds}. The DAS collects driving data continuously while the vehicle is operating, with data collection frequency being 10 Hz \cite{zhu2018human, zhu2018modeling}.

\subsection{Car-Following Event Extraction}
Car-following events were extracted from the dataset using the same criteria from previous studies \cite{zhu2018human, wang2017capturing, zhao2017trafficnet}. The specific criteria are shown below:

\begin{itemize}
    \item Front target's identification number remained constant: indicating the subject vehicle was following the same lead vehicle;
    \item Lateral distance between the two vehicles < 2.5m: making sure that they drove in the same lane; and
    \item Period duration > 15s: guaranteeing that the car-following event contained enough data to be analyzed.
\end{itemize}




  
  A total of 112,597 car-following events were extracted and utilized in this study. The train, validation, and test splits are 70\% (78,817), 15\% (16,890), and 15\% (16,890). The train data are used for model training; the validation data are used for model selection; and the testing data are used for performance reporting.

\subsection{Baselines}
Three baseline models are used as a comparison to demonstrate the performance of our proposed model:
\begin{itemize}
    \item \textbf{IDM model}: according to Zhu et al. \cite{zhu2018modeling}, IDM is a traditional car-following model with the best behavior prediction performance compared to other traditional models. We trained an IDM model using the training dataset. The genetic algorithm was used to find the best IDM parameter set to minimize the validation dataset MSE (see \cite{zhu2018modeling} for IDM calibration details). 
    \item \textbf{Fully connected neural network (NN)}: a three-layer feedforward neural network with ReLU activations is used to predict future FV speed based on the same inputs to the decoder part of the TransFollower model. The hidden layer dimension of the NN is 256 (same as the hidden vector dimension in TransFollower). The first layer transformer inputs in the shape of $\mathbb{R}^{120 \times 2}$ to $\mathbb{R}^{120 \times 256}$. The second layer's outputs are still in the shape of $\mathbb{R}^{120 \times 256}$, which are then transformed by the third layer to the shape of $\mathbb{R}^{120 \times 1}$. These final outputs are considered as predicted future FV speed. The network is then optimized with MSE loss. 
    \item \textbf{LSTM based model}: this baseline model shares a similar architecture with the TransFollower model. It has an encoder and a decoder part, both formed by a stack of 4-layer LSTM models (hidden dimension = 256, dropout probability = 0.4). The inputs to the encoder and decoder are the same as the TransFollower model. The Adam optimizer with a learning rate = 0.001 was used to optimize all models. A batch size of 256 was used. 
\end{itemize}

\subsection{Train and Test Details}
All the car-following events are trimmed to be in the length of 15 seconds. The first 4 seconds (40 time steps since data sampling interval = 0.1 seconds) are considered as historical driving context, and the remaining 11 seconds of FV speed data are the target we want to predict. For the inputs to the decoder part, one second (10 time steps) of historical data are used as an initial token to the decoder. 

\section{Results}
\subsection{MSE in Testing Dataset}

\begin{figure}[!h]
    \centering
    \subfloat[Speed]{\includegraphics[width=0.8\textwidth]{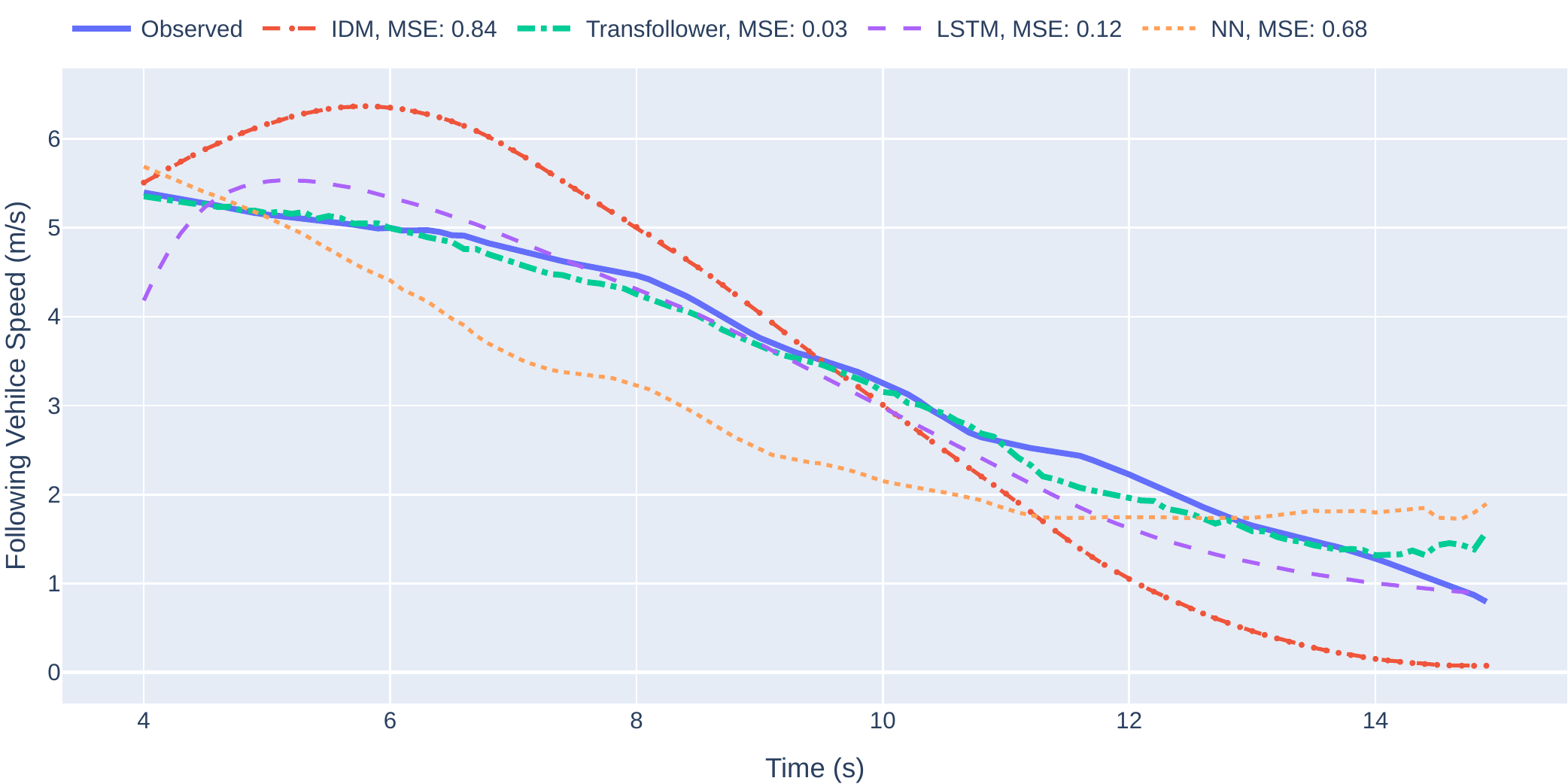}}
    \hfill
    \subfloat[Spacing]{\includegraphics[width=0.8\textwidth]{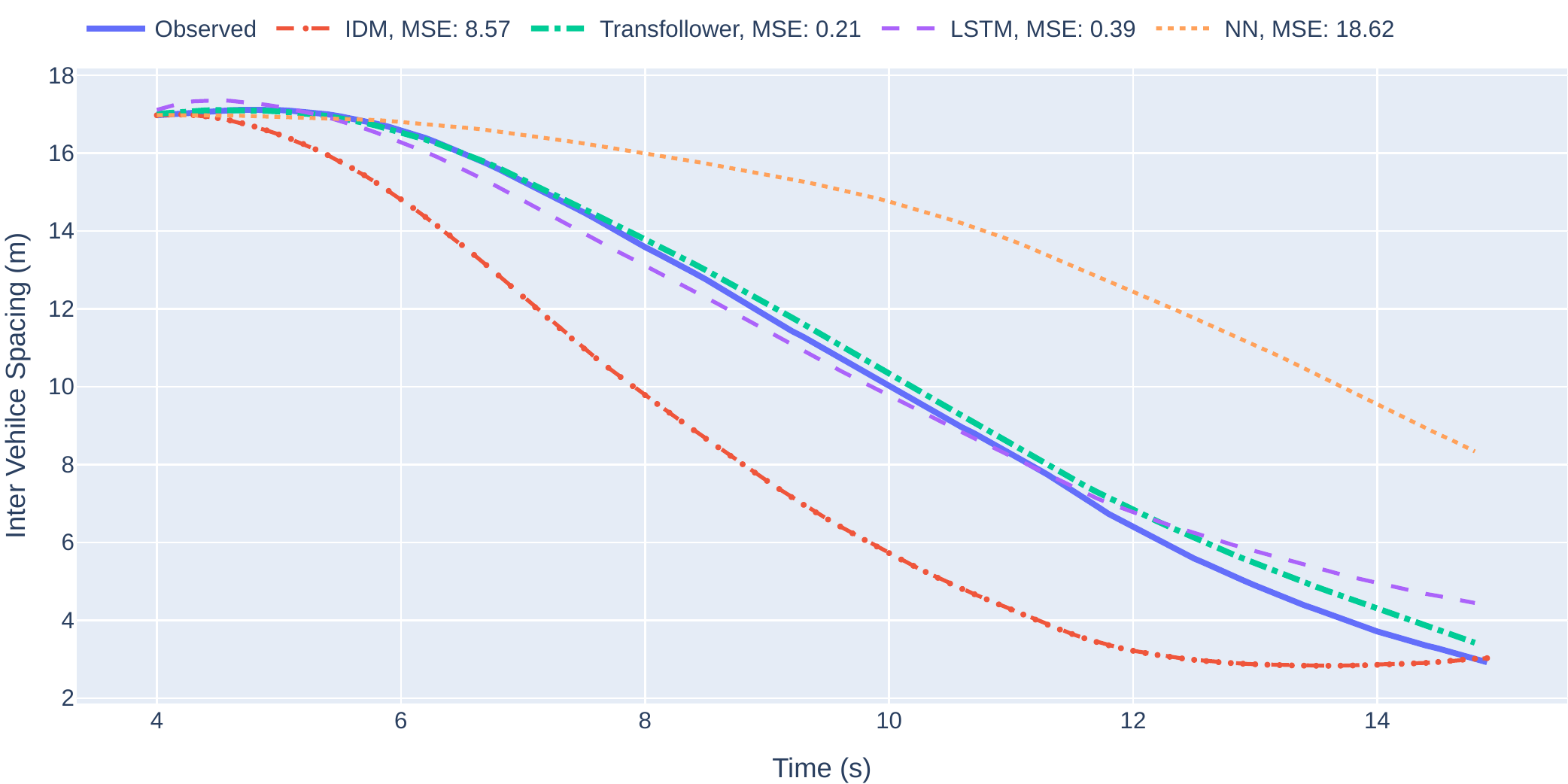}}
    \caption{Observed and simulated car-following trajectories for different models (example \#1).}
    \label{fig:traj1}
\end{figure}

\begin{figure}[!h]
    \centering
    \subfloat[Speed]{\includegraphics[width=0.8\textwidth]{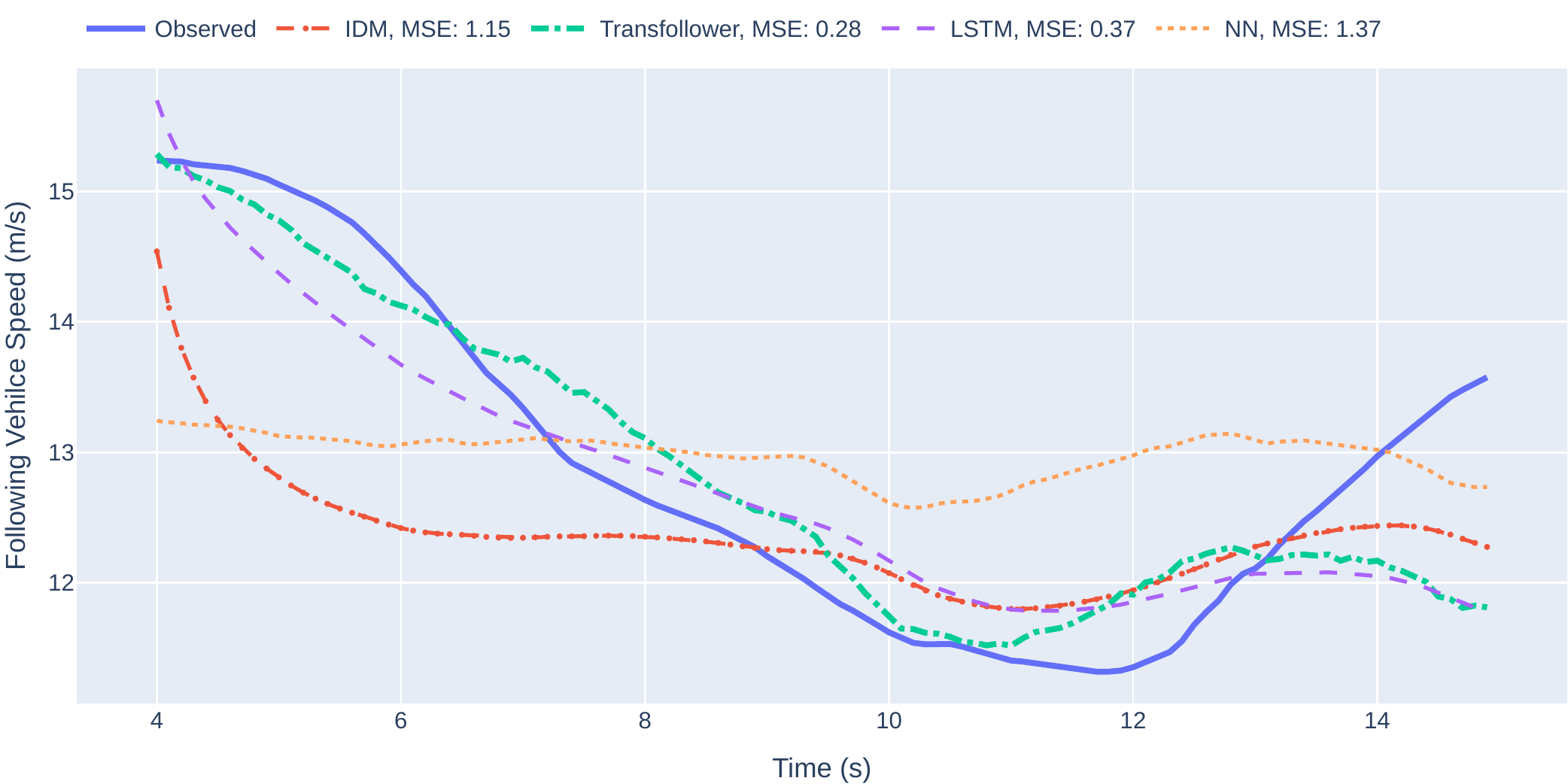}}
    \hfill
    \subfloat[Spacing]{\includegraphics[width=0.8\textwidth]{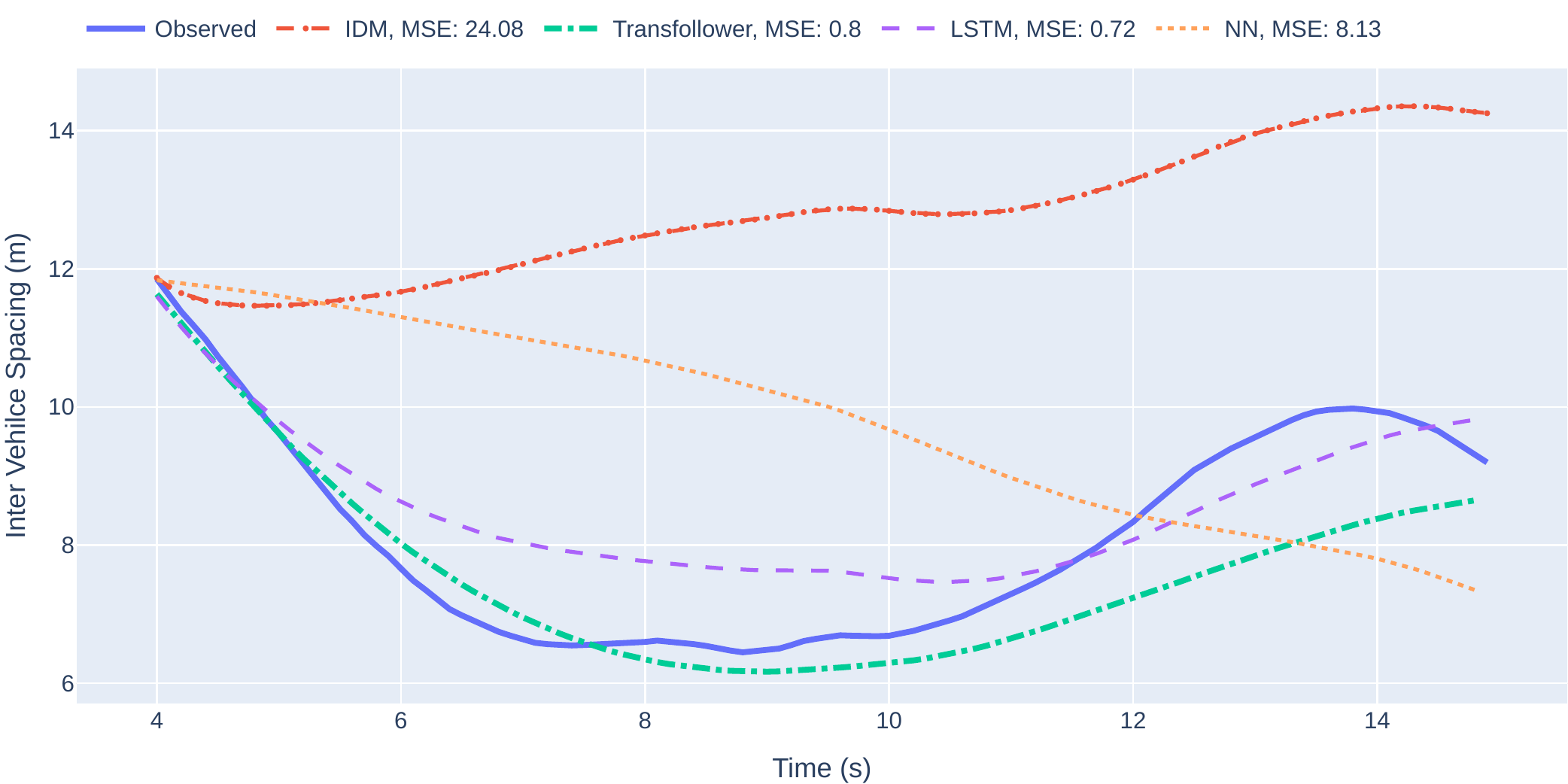}}
    \caption{Observed and simulated car-following trajectories for different models (example \#2).}
    \label{fig:traj2}
\end{figure}

   
The sum of MSE for both spacing and FV speed in the testing dataset is used to measure the performance of models, as shown in Table \ref{tab:results}. Our proposed model, TransFollower, outperforms the IDM, NN-, and LSTM- based models by a large margin. Fig. \ref{fig:traj1} and \ref{fig:traj2} show two examples of car-following trajectories predicted by different models. In these two examples, the predicted speed and spacing trajectories by the TransFollower model are closer to the ground-truth trajectories compared to other models. 

\begin{table}[!h]
\caption{Model performance comparison.}
\centering
\begin{tabular}{@{}lllll@{}}
\toprule
Model           & Testing MSE   \\ \midrule
IDM     & 22.5     \\
NN & 50.6      \\
LSTM           & 31.8     \\ 
\textbf{TransFollower (ours)} & \textbf{8.07}\\ \bottomrule
\end{tabular}
\label{tab:results}
\end{table}


\subsection{Visualization of Attentions}
To understand what is learned in the proposed attention-based model, we visualized different kinds of attention for some sample car-following events. Attention in this paper means a distribution of weights applied to different locations of an input. A higher attention value at a certain position means the input value at this position has a higher impact on the output. In our proposed TransFollower model, there are 3 types of attention:
\begin{itemize}
    \item Encoder self-attention: reflecting how the mixed representation of historical driving context generated by the encoder pays attention to the encoder inputs, which include FV speed, relative speed, and spacing data;
    \item Decoder self-attention: reflecting how decoder hidden outputs pay attention to the input LV speed profile;
    \item Encoder-decoder cross-attention: reflecting how the decoder pays attention to the encoder representation of historical driving context, to generate final prediction results. 
\end{itemize}

\subsubsection{Encoder Self-Attention}
\begin{figure}[!h]
    \centering
    \subfloat[First Layer]{\includegraphics[width=1\textwidth]{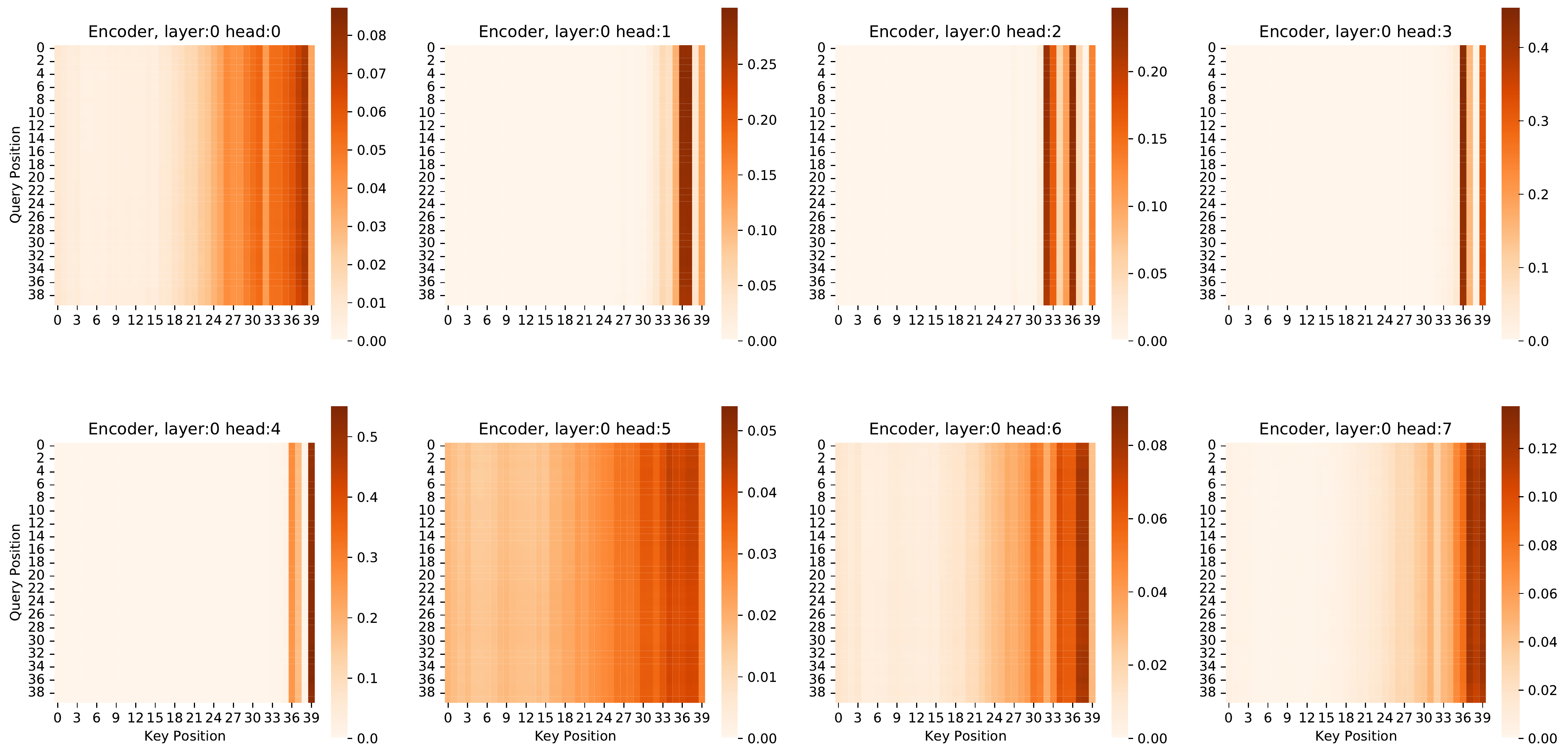}}
    \hfill
    \subfloat[Second Layer]{\includegraphics[width=1\textwidth]{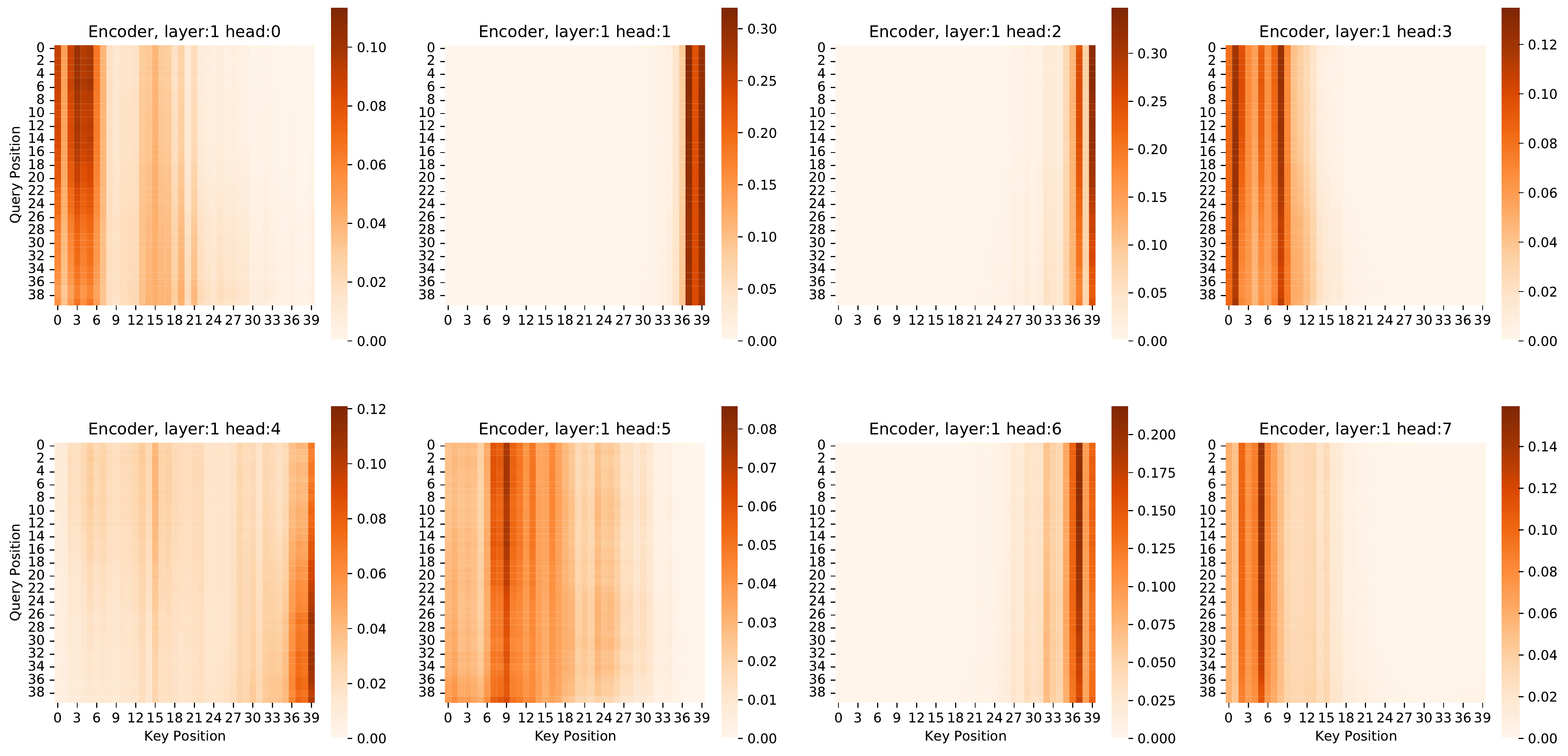}}
    \caption{Encoder self-attention visualization for one example car-following event.}
    \label{fig:enc_attn}
\end{figure} 
Fig. \ref{fig:enc_attn} shows the encoder multi-head self-attention heatmaps for one sample car-following event. Each row in the heatmap sums up to 1, and the values represent the distribution of weights to different positions of the key vector. Since we have two layers for the encoder, the attention heatmaps at different layers are shown separately. 

For the first layer (Fig. \ref{fig:enc_attn}-a), the first interesting finding is that different attention heads have different patterns of attention and thus formulate their output differently. Another finding is that most attention heads pay more attention to the most recent historical driving inputs (position 30 to 40). This is reasonable because the inputs to the first layer still maintain the general time sequencing of raw inputs, and future driving will be more affected by the recent driving context. 

At the second layer, although there are still some heads that pay more attention to the recent inputs, this pattern does not dominate as in the first layer. E.g., attention heads 0, 3, and 7 pay more attention to the beginning part of the input sequence. One possible reason is that through layer 1 of the encoder part, most positions at the second layer’s inputs already contain much information for the most recent driving context. This also means that at the second layer, the hidden vectors are more mixed in time sequencing than the first layer. 

\subsubsection{Decoder Self-Attention}
\begin{figure}[h]
    \centering
    \includegraphics[width=1\textwidth]{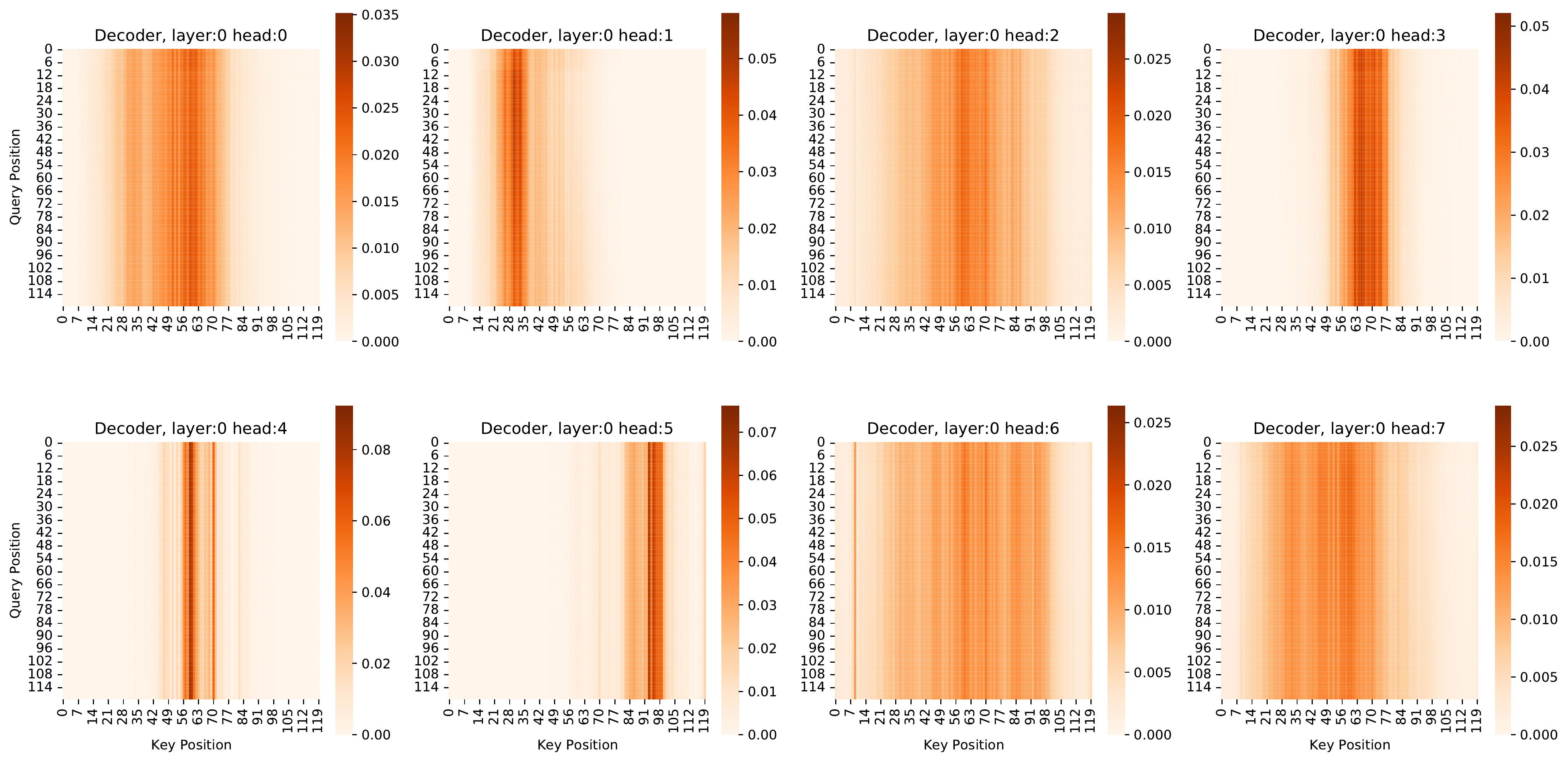}
    \caption{Decoder self-attention visualization for one example car-following event.}
    \label{fig:dec_attn}
\end{figure}

Fig. \ref{fig:dec_attn} shows the decoder multi-head self-attention heatmaps for the sample car-following event as in the encoder self-attention part. For this part, we do not see a dominant pattern that pays more attention to the ending part of the input sequence, as in the first layer of the encoder. Instead, different heads will pay attention to different regions of the input sequence. This is because the decoder aims to predict a multi-step speed profile for the FV, and predicted speed at different time steps will depend on different parts of the LV speed profile.  

\begin{figure}[h]
    \centering
    \includegraphics[width=0.8\textwidth]{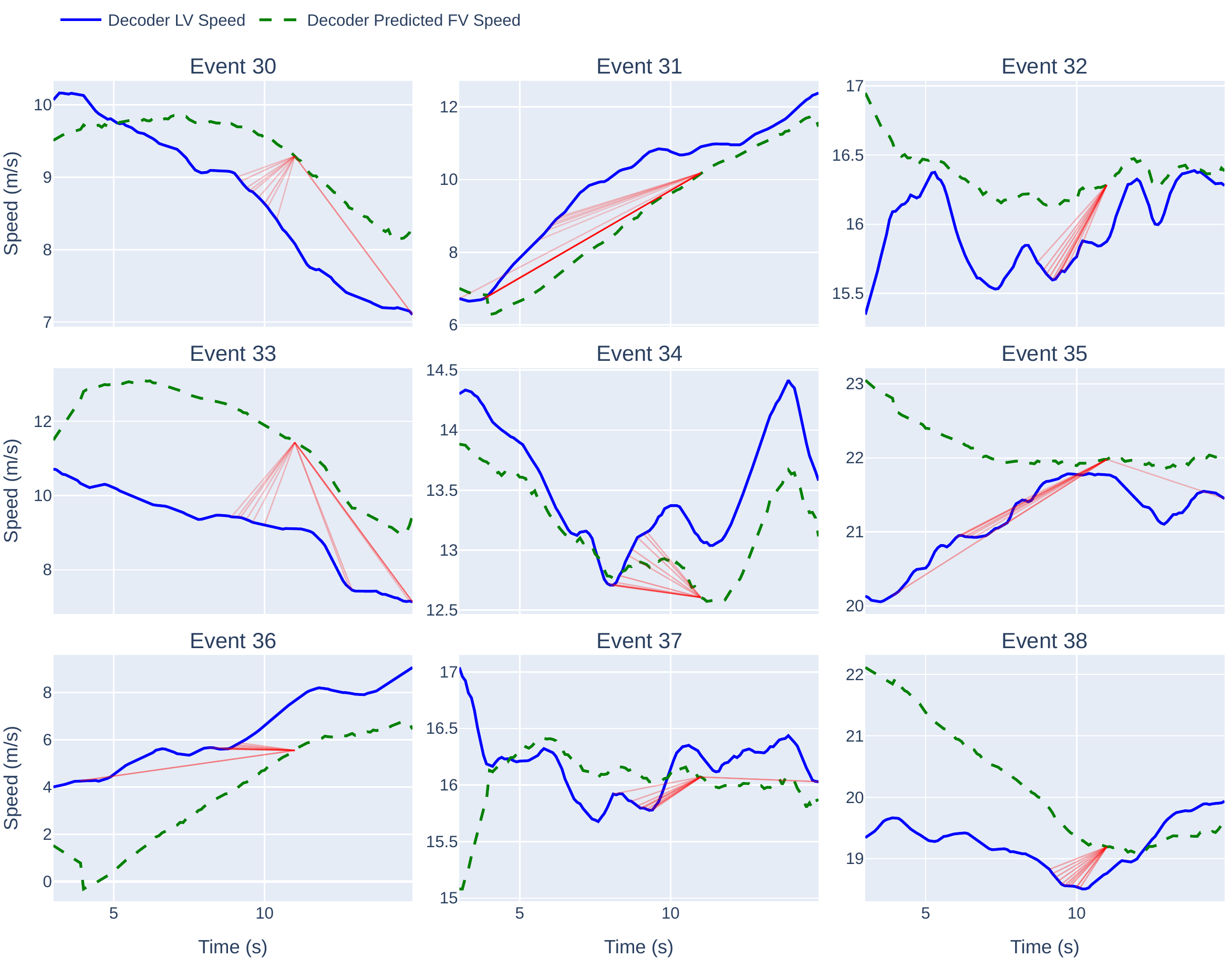}
    \caption{Average multi-head attention at one position of the decoder predicted FV speed profile, for multiple car-following events. For better clear visualization, only the top 10 attention lines are presented, with the line opacity positively correlated to the attention weight.}
    \label{fig:one_point_dec_attention}
\end{figure}

Fig. \ref{fig:one_point_dec_attention} shows how the predicted FV speed at one position (11s) relates to the inputted LV speed profile. We present the top 10 attention lines among the total 120 lines to see which regions of the input LV speed profile have the biggest impacts on the FV speed prediction. For most events, the predicted FV speed is more related to the LV speed profile during the past 1 to 2 seconds. This is because the most recent LV speed patterns will convey more information for FV speed prediction. Another interesting finding is that peak points in the LV speed profile have a larger impact on the predicted FV speed (e.g., events 32, 34, and 37). One possible reason for this is that peak points contain information about the acceleration changes of LV speeds, which is important for speed control. Since we do not impose any masks on the decoder attention masks, we can see some attention lines pointing to the future LV speed positions. The benefit of this is that we can predict the FV speed in an integrated and coherent way by referencing both past and future LV speed information. 

\subsubsection{Encoder-Decoder Cross-Attention}
\begin{figure}[h]
    \centering
    \includegraphics[width=0.8\textwidth]{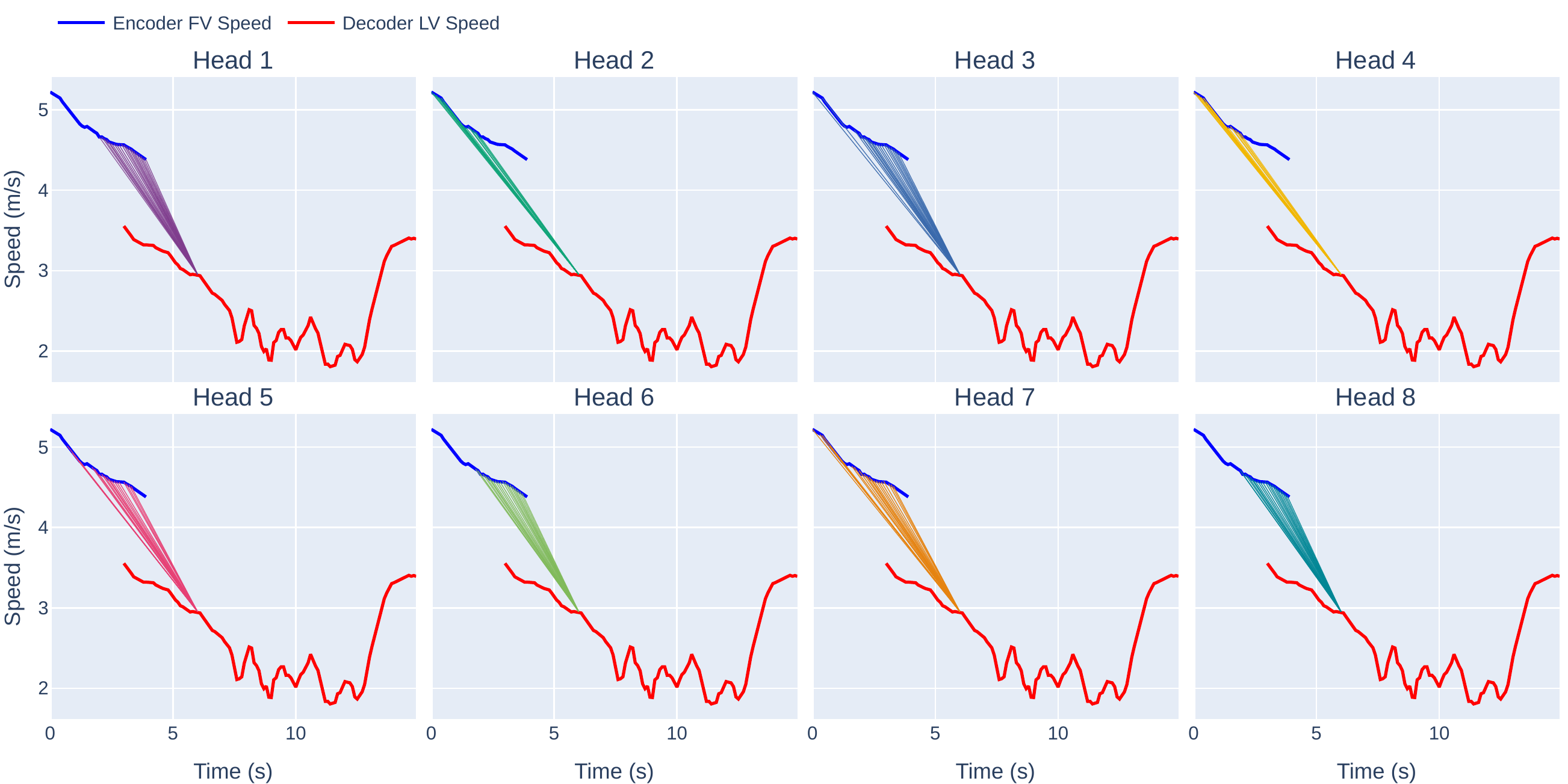}
    \caption{Multi-head cross-attention between encoder FV speed and Decoder LV speed for one sample car-following event.}
    \label{fig:one_point_cross_attention}
\end{figure}

Fig. \ref{fig:one_point_cross_attention} shows the multi-head cross attention for one sample car-following event. We look at how one point of LV speed profile at the decoder attends to the encoder FV speed profile. Again, only the top 20 (among the total of 40) attention lines are shown here. For cross attention, different heads will pay attention to different parts of encoder outputs.

\section{Summary and Conclusion}
This study aims to address the two major challenges in car-following behavior modeling: 1) how to model the temporal dependency of future actions on historical driving context, and 2) how to predict long-sequence car-following trajectories accurately without introducing compounding errors accumulated through multiple steps. Inspired by the recent success of attention-based Transformer models in language modeling, we proposed an encoder-decoder architecture-based car-following model with Transformer blocks as the backbone of both encoder and decoder. The encoder processes historical speed and spacing data and builds a mixed representation of historical driving context using multi-head self-attention. The decoder takes the future FV speed profile as input and outputs the predicted future FV speed profile in a generative way (instead of an auto-regressive way, avoiding compounding errors). Through cross-attention between encoder and decoder, the decoder learns to build a connection between historical driving and future LV speed, based on which a prediction of future FV speed can be obtained. Based on real-world car-following events from the SH-NDS, we demonstrated that the model has superior performance on long-sequence car-following trajectory prediction, outperforming traditional car-following models and some recent data-driven models. To show how the model derives its predictions, we also visualized the self-attention and cross-attention heatmaps for model understanding. 

The conclusions of this study can be summarized as:
\begin{itemize}
    \item The proposed Transformer-based model has superior performance on historical dependency parsing and long-sequence trajectory prediction. This is supported by its lowest MSE on spacing and speed in the testing dataset. 
    \item Generative way of prediction outperforms the auto-regressive way for long-sequence car-following trajectory prediction.
    \item For both self-attention and cross attention in the model, different attention heads have different patterns of weights distribution and thus formulate their output differently.
    \item In the lower layer of the encoder part, most attention heads pay more attention to the most recent historical driving inputs (1 to 2 seconds before the current time step), indicating that future driving will be more affected by recent driving context.
    \item The predicted FV speed is more related to the LV speed changes during the past 1~2 seconds, suggesting that the most recent LV speed patterns convey more information for FV speed prediction. Also, peak points in the LV speed profile have a larger impact on the predicted FV speed since they contain information about the acceleration changes of LV speeds.
    
\end{itemize}

\section{Acknowledgements}
The authors would like to thank the Pacific Northwest Transportation Consortium (PacTrans) at Regional University Transportation Center (UTC) for Federal Region 10, for funding this research. 

\newpage

\bibliographystyle{trb}
\bibliography{trb_template}
\end{document}